\documentclass[letterpaper, 10 pt, conference]{ieeeconf}  

\usepackage{amsmath,amsfonts}
\usepackage{array}
\usepackage{algorithm}
\usepackage{algorithmic}
\usepackage{amsmath}
\usepackage{cite}
\usepackage{graphicx}

\usepackage{xcolor}
\usepackage{capt-of}
\usepackage{booktabs}
\usepackage{multirow}

\usepackage{tikz}
\usetikzlibrary{arrows.meta,patterns,calc}
\usepackage{xcolor}
\usepackage[hidelinks]{hyperref}

\makeatletter
\renewcommand{\@IEEEfigurecaptionsepspace}{\vskip 3pt\relax}
\makeatother
\newcommand{\PHILfigurebottomspace}{\par\vspace{-6pt}}

\IEEEoverridecommandlockouts                              

\title{\LARGE \bf
Correct Robots Before They Make Mistakes: Proactive Human-in-the-Loop Intervention via Preference Learning
}

\author{Xinkai Jiang$^{1,*}$, Jing Jiang$^{1}$, Hongyi Zhou$^{1}$,
Pankhuri Vanjani$^{1}$, Ahmad Baki$^{1}$,
Jakub Suliga$^{1}$, Ge Li$^{1}$ \\ Zhuoyue Li$^{1}$, Gerhard Neumann$^{1}$, Rudolf Lioutikov$^{1,2}$%
\thanks{$^{1}$Karlsruhe Institute of Technology (KIT), Germany.}%
\thanks{$^{2}$Robotics Institute Germany (RIG).}%
}

\begin{document}

\IEEEaftertitletext{%
  \noindent%
    \begin{minipage}{\textwidth}
      \centering
      \includegraphics[width=\linewidth]{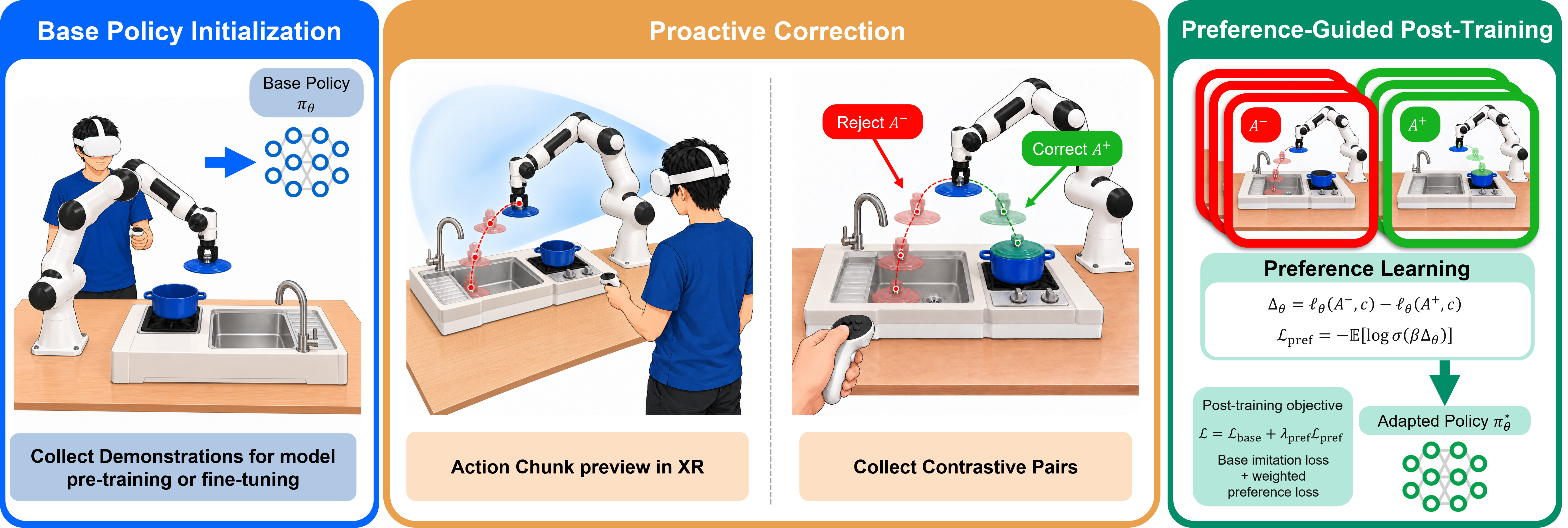}
      \captionof{figure}{Overview of the Proactive Human-in-the-Loop (PHIL) framework.
      Demonstrations collected through any suitable interface are used to pre-train or fine-tune a base policy.
      During rollouts, the operator previews predicted future end-effector trajectories in XR and takes control before an anticipated error.
      Policy queries continue on a fixed schedule during human takeover.
      Preference pairs consist of a rejected policy action chunk $A^-$ (red) and the next $H$ executed actions $A^+$ (green), both starting from the same query state.
      Post-training combines imitation learning on the base demonstrations with a preference objective favoring $A^+$ over $A^-$.}
      \label{fig:teaser}
    \end{minipage}
  \endgraf\vspace{4pt}%
}

\maketitle
\thispagestyle{empty}
\pagestyle{empty}




\begin{abstract}

This paper presents Proactive Human-in-the-Loop (PHIL), a framework for proactive policy correction and preference-based post-training.
PHIL enables operators to preview policy-predicted future motion and intervene before anticipated undesirable actions.
During takeover, PHIL continues querying the policy and displaying its predicted trajectories in the physical workspace through an extended reality (XR) interface. 
PHIL pairs rejected policy predictions with corrective action chunks subsequently executed from the same physical state over the same horizon, and preference-based post-training encourages the policy to favor these corrections.
This paper evaluates PHIL on four real-world manipulation tasks using a diffusion policy and a pretrained vision-language-action policy.
With five correction episodes per condition, PHIL increases mean success rates across the evaluated tasks and demonstration budgets from 44.0\% to 70.3\% for the diffusion policy and from 57.8\% to 76.9\% for X-VLA, compared with HG-DAgger.
Separately, policies trained with proactive corrections show fewer unnecessary recovery behaviors in successful episodes.
Project page: \url{https://intuitive-robots.github.io/PHIL-page/}.

\end{abstract}


\section{Introduction}
\label{sec:introduction}

Robot policies, including visuomotor policies and vision-language-action (VLA) models, have shown strong performance on a broad range of manipulation tasks.
However, policies trained from offline demonstrations may still encounter states that are beyond the coverage of their training data.
In such situations, the policy may produce undesirable motions or fail to complete the task.
Improving performance in such cases requires additional task-specific data.

Human-in-the-loop imitation learning addresses this gap by collecting corrective data during policy rollouts.
During these rollouts, the policy may visit states that are absent from, or underrepresented in the offline demonstrations.
Most existing human-in-the-loop methods collect correction data reactively.
In reactive correction, the operator decides whether to intervene based on the robot's currently observed execution, without a preview of the policy's predicted future motion.
It is straightforward to deploy and allows corrective data to be collected during policy rollouts.
The resulting recovery actions can also be valuable when the robot must recover from an error and retry a task~\cite{Hu2025RaC}.
This approach has enabled effective policy refinement through interactive data aggregation, remote teleoperation, and corrective residual control~\cite{hg-dagger,Mandlekar2020RemoteTeleop,xu2025crdagger}.
When reactive corrections begin after an error has occurred, they include recovery actions in addition to actions that advance the task.
The proportion of reactive correction data in the dataset can affect how often recovery motions appear in the policy output.
A higher proportion increases the likelihood of producing recovery motions even when no recovery is needed.

These limitations motivate proactive correction, in which the operator reviews the policy's proposed motion and provides corrections before an anticipated error occurs.
By visualizing the policy's predicted future motion, proactive correction enables the operator to identify and correct potential errors before the robot enters safety-critical states.
This predictive preview reduces the operator's cognitive burden of continuously anticipating the policy's behavior.
It can also reduce the need for repeated reactive corrections as an error unfolds.
Although proactive correction shows considerable promise, it has received limited attention.
Existing work has considered simulated tasks in autonomous driving and robotic manipulation~\cite{Cai2025PPL}, while its application to complex physical robot manipulation remains unexplored.
Extending proactive correction to such tasks requires an interface that allows the operator to intuitively assess the robot's proposed three-dimensional motion before an anticipated error occurs.
Although a screen-based preview can overlay this motion on a camera view, judging its depth and spatial orientation from a two-dimensional display can be difficult.
To address this challenge, we propose PHIL, a proactive human-in-the-loop imitation learning framework that uses an Extended Reality (XR) interface to visualize proposed policy action chunks in the physical workspace.
Each action chunk is converted into a sequence of three-dimensional end-effector poses and spatially registered with the physical workspace.
The operator can therefore inspect upcoming motion in three dimensions within its execution context while the robot executes the current action chunk.

When the preview indicates that the proposed motion may lead to an error, the operator takes control and provides corrections before the anticipated undesirable action occurs.
During the takeover, PHIL continues to query the policy at the unchanged query schedule and display updated action chunks in the XR headset.
PHIL constructs contrastive pairs from the displayed policy action chunks and the corresponding executed corrective robot trajectories.
For each retained policy query during intervention, the displayed action chunk is recorded as rejected, denoted by \(A_t^-\).
The next \(H\) executed robot actions starting at that query form the corresponding corrective action chunk \(A_t^+\), which may include policy-controlled actions after release.
Both action chunks in each pair start from the same physical state.
When the operator judges the most recently displayed action chunk to be acceptable, they release control and the robot returns to policy control.
After data collection, PHIL uses the contrastive pairs to fine-tune the policy, encouraging it to favor executed corrective action chunks \(A_t^+\) over rejected policy action chunks \(A_t^-\) from the same physical state.
We evaluate PHIL on four real-world manipulation tasks, with experiments covering both diffusion and pretrained VLA policies.
Compared with HG-DAgger, PHIL increases the average success rate from 44.0\% to 70.3\% for the diffusion policy and from 57.8\% to 76.9\% for X-VLA.
Proactive correction data reduce unnecessary recovery behaviors approximately tenfold compared with reactive correction data.

In summary, this work makes the following contributions:
\begin{itemize}
    \item We introduce PHIL, a proactive human-in-the-loop framework that leverages XR to visualize proposed robot motions in the physical workspace, enabling human corrections before anticipated errors occur.
    \item We extract rejected--corrected action-chunk pairs anchored at the same physical state during human takeover and use them for preference-based post-training to favor executed corrective trajectories over rejected policy actions.
    \item We evaluate PHIL on real-world manipulation tasks with diffusion and pretrained VLA policies, demonstrating improved task success and showing that proactive correction data reduce unnecessary recovery behaviors.
\end{itemize}

\section{Related Work}

\subsection{Human-in-the-Loop Robot Learning}

Prior human-in-the-loop robot learning methods collect additional training data through human takeovers during policy rollouts, using assistance requests and control switching to reduce intervention burden~\cite{hg-dagger,hoque2022thriftydagger,hoque2021lazydagger,Hoque2024IntervenGen}.
Recent methods extend interactive learning to generative visuomotor and vision-language-action policies through uncertainty-guided intervention, corrective post-training, and online adaptation~\cite{He2025Uncertainty,Han2025DexHiL,Murray2026FlowDAgger,Welte2026FlowCorrect}.
These methods address supervision allocation and policy learning.
PHIL informs intervention decisions through an XR preview of proposed motion in the physical workspace, enabling takeover before anticipated errors.

\subsection{Corrective Imitation Learning}

Corrective imitation learning mitigates distribution shift in behavior cloning by collecting supervision at states encountered during policy rollouts~\cite{ross2011reduction}.
For action-chunking visuomotor policies, which predict short action sequences~\cite{reuss2023goal}, an unsuitable motion may be partly executed before its consequences become apparent to the operator.
PPL visualizes predicted future trajectories to support proactive intervention and propagates intervention-derived preferences to predicted future states~\cite{Cai2025PPL}.
SDP uses paired undesired and corrective action chunks to construct sets of desired actions for diffusion-policy training~\cite{Li2026SDP}.
PHIL uses proactive correction for preference-based post-training of diffusion and pretrained VLA policies in real-world manipulation.
During human takeover, it pairs each retained, unexecuted policy prediction with the subsequent executed corrective action chunk from the same physical state and spanning the same temporal horizon.

\subsection{Extended Reality for Robotics}

Extended Reality (XR) supports spatial visualization and three-dimensional interaction for robot learning.
AR-based systems enable robot-free demonstration collection~\cite{ar2-d2-pmlr-v229-duan23a,eve} and provide visual feedback for collecting demonstrations~\cite{arcade,chen2024arcap,jiang2024comprehensive}.
Other interfaces support immersive teleoperation and demonstration collection~\cite{holodex,opentelevision}, interactive skill maintenance~\cite{arclfd}, and robot interaction across simulated and physical environments~\cite{jiang2025iris}.
VR-DAgger further combines immersive interaction with uncertainty-guided selection and correction of policy rollout segments~\cite{Zurbrugg2026VRDagger}.
PHIL renders the future end-effector trajectory implied by a policy action chunk in registration with the physical workspace, allowing the operator to assess upcoming motion relative to the robot and surrounding objects during execution and intervene proactively.

\section{Method}
\label{sec:method}

\begin{figure}[t]
\centering
\includegraphics[width=\columnwidth]{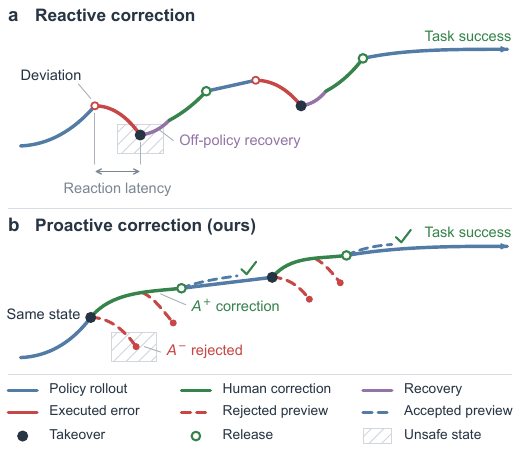}
\caption{Reactive versus proactive correction under equal correction budgets (two interventions per panel; only the first annotated).
Solid and dashed lines denote executed robot motion and previewed policy continuations, respectively.
Top: in these examples, reactive intervention follows observed errors and requires recovery.
Bottom: XR previews enable intervention before anticipated errors.
During correction, rejected predictions $A^{-}$ from scheduled queries are paired with subsequent executed corrective trajectories $A^{+}$ from the same query state.
Takeover and release leave the query schedule unchanged; control returns to the policy when the preview is acceptable.}
\label{fig:proactive_correction}
\PHILfigurebottomspace
\end{figure}

PHIL post-trains an action-chunking robot policy using proactive corrections collected before an anticipated undesirable action occurs.
As illustrated in Fig.~\ref{fig:proactive_correction}, the operator inspects the policy's predicted future motion during execution.
If an error is anticipated, the operator takes control of the robot to provide corrective actions.
The policy continues to generate predictions on its fixed query schedule during the takeover.
Retained predictions are paired with subsequent executed corrective action chunks from the same query state and used for preference-based post-training.

\subsection{Problem Formulation}

Consider an action-chunking policy $\pi_\theta$ operating in discrete time.
At time $t$, its context is $c_t=(o_t,q_t,\ell_t)$, where $o_t$ contains the current visual observations, $q_t$ is the robot proprioceptive state, and $\ell_t$ is the task instruction.
This formulation covers both visuomotor and vision-language-action policies.
Given $c_t$, the policy predicts an action chunk of horizon $H$,
\begin{equation}
A_t
=
\left(a_t^{(0)},a_t^{(1)},\ldots,a_t^{(H-1)}\right)
\sim \pi_\theta(\cdot\mid c_t),
\label{eq:action_chunk}
\end{equation}
where $a_t^{(k)}\in\mathcal A$ is the action at offset $k$ in the proposal generated at time $t$.
Each action chunk is converted into a trajectory of absolute Cartesian end-effector poses for the operator to preview and assess.

The policy is queried every $B$ control steps, with $B\leq H$.
Let
\begin{equation}
\mathcal T_{\mathrm{query}}
=
\{t_0+nB\mid n=0,1,2,\ldots\}
\label{eq:query_schedule}
\end{equation}
denote the fixed query schedule, where $t_0$ is the first query time in a rollout.
Policy queries occur only at the scheduled times, both during policy control and during human intervention.
When the robot remains under policy control between consecutive queries, it executes the first $B$ actions of the predicted chunk.
The full $H$-step chunk is available for the operator to preview.

\subsection{Collecting Preference Pairs through Proactive Correction}
\label{sec:proactive_collection}

Correction data are collected using a base policy $\pi_{\theta_0}$ whose parameters remain fixed throughout collection.
Let $\tilde a_t$ denote the robot action actually executed at time $t$, whether commanded by the operator or the policy.
Suppose the operator takes over at time $r$ and controls the robot for $M$ steps.
The policy continues to run in shadow mode at the scheduled query times
\begin{equation}
\mathcal T_{\mathrm{shadow}}
=
\{t\in\mathcal T_{\mathrm{query}}\mid r\leq t<r+M\}.
\label{eq:shadow_queries}
\end{equation}
At each $t\in\mathcal T_{\mathrm{shadow}}$, it generates $A_t\sim\pi_{\theta_0}(\cdot\mid c_t)$.
These predictions are displayed and recorded but are not executed while the operator controls the robot.

PHIL infers rejection from continued human intervention and treats the shadow predictions as candidates for negative samples.
Predictions accepted at release are excluded from negative samples.
The robot then returns to policy control, including the interval before the next scheduled query.

A shadow prediction collected at time $t$ during an intervention is paired with the next $H$ executed robot actions to form a candidate preference pair:
\begin{equation}
\begin{aligned}
A_t^- = A_t, \quad
A_t^+ = \left(\tilde a_t,\ldots,\tilde a_{t+H-1}\right).
\end{aligned}
\label{eq:preference_chunks}
\end{equation}
Both chunks originate from the same query state and span the same $H$-step horizon.
The executed corrective continuation is treated as preferred to the rejected prediction, yielding the label $A_t^+ \succ A_t^-$.

\begin{figure}[t]
\centering
\includegraphics[width=\columnwidth]{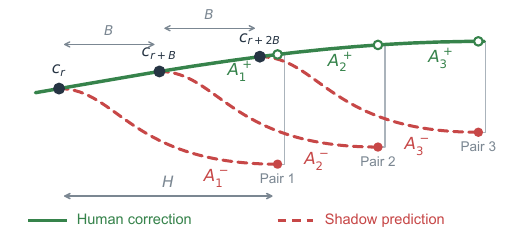}
\caption{Preference pairs from fixed $B$-step queries during human intervention.
Each retained rejected $H$-step prediction $A_t^-$ pairs with the next $H$ executed actions $A_t^+$, yielding $A_t^+\succ A_t^-$.
Here, takeover coincides with a query and positive windows fall entirely within the human intervention period; generally, windows may include policy actions after release.
Windows overlap when $B<H$.
Takeover and release do not reset queries.
Only complete pairs are retained; the latest prediction displayed at release is accepted and excluded from negatives.}
\label{fig:proactive_preference_pairs}
\PHILfigurebottomspace
\end{figure}

Aggregating the complete pairs across all interventions and rollouts gives the preference dataset
\begin{equation}
\mathcal D_{\mathrm{pref}}
=
\left\{(c_i,A_i^+,A_i^-)\right\}_{i=1}^{N},
\label{eq:preference_dataset}
\end{equation}
where $i$ indexes the $N$ retained pairs.
Fig.~\ref{fig:proactive_preference_pairs} illustrates the construction of multiple pairs during an intervention.

\subsection{Preference Post-Training with Imitation Learning}
\label{sec:preference_posttraining}

After data collection, the trainable policy $\pi_\theta$ is initialized from the base-policy $\pi_{\mathrm{base}}$.
Let $\ell_\theta(A,c)$ denote the policy's native imitation loss for action chunk $A$ under context $c$, averaged over all $H$ action positions.
Depending on the policy, this can be a negative log-likelihood, diffusion denoising, or flow-matching loss.
All retained chunks are complete, so no temporal padding or masking is needed for the preference pairs.
The pairwise objective uses the imitation loss as a surrogate for comparing the two action chunks.

Post-training retains imitation learning on the original base demonstrations:
\begin{equation}
\mathcal L_{\mathrm{base}}
=
\mathbb E_{(c,A)\sim\mathcal D_{\mathrm{base}}}
\left[\ell_\theta(A,c)\right].
\label{eq:base_imitation_loss}
\end{equation}
For each preference tuple $(c,A^+,A^-)$, the loss difference is defined as
\begin{equation}
\Delta_\theta
=
\ell_\theta(A^-,c)-\ell_\theta(A^+,c).
\label{eq:preference_margin}
\end{equation}
A positive $\Delta_\theta$ indicates that the preferred corrective chunk receives a lower imitation loss than the rejected prediction.
The pairwise logistic objective encourages this ordering:
\begin{equation}
\mathcal L_{\mathrm{pref}}
=
-\mathbb E_{(c,A^+,A^-)\sim\mathcal D_{\mathrm{pref}}}
\left[\log\sigma\left(\beta\Delta_\theta\right)\right],
\label{eq:preference_loss}
\end{equation}
where $\sigma$ is the logistic sigmoid and $\beta>0$ controls the sharpness of the comparison.
This objective encourages a lower loss for the corrective chunk relative to the rejected prediction, with both losses averaged over the same $H$-step horizon.

The complete post-training objective is
\begin{equation}
\mathcal L
=
\mathcal L_{\mathrm{base}}
+
\lambda_{\mathrm{pref}}\mathcal L_{\mathrm{pref}},
\label{eq:total_loss}
\end{equation}
where $\mathcal L_{\mathrm{base}}$ is the imitation loss on the original base demonstrations defined in Eq.~\eqref{eq:base_imitation_loss}, and $\lambda_{\mathrm{pref}}$ weights the preference loss relative to this term.
The pair construction and pairwise objective are shared across policy families, while the native imitation loss $\ell_\theta$ depends on the underlying policy.

\section{System Overview}

\subsection{Teleoperation System for Interactive Learning}

The PHIL teleoperation system uses a Meta Quest~3 headset and motion controller to support two interactive data-collection modes: demonstration mode and correction mode.
In demonstration mode, the operator directly controls the robot to collect expert trajectories for policy pre-training or fine-tuning.
In correction mode, the system allows smooth switching between policy control and human intervention.
The operator can take over by holding the hand trigger to provide a correction, and releasing the trigger returns control to the policy.
During corrective data collection, the system records both the executed robot trajectory and the policy outputs.

The Meta Quest~3 controller provides its absolute Cartesian pose for teleoperation.
When the operator presses the hand trigger, the system aligns this pose with the current end-effector pose and enables teleoperation until the trigger is released.
Relative controller motions are then mapped to desired end-effector poses, avoiding control discontinuities when teleoperation starts or resumes.
The index trigger is used to close and open the gripper.

\begin{figure*}[!t]
    \centering
    \includegraphics[width=\textwidth]{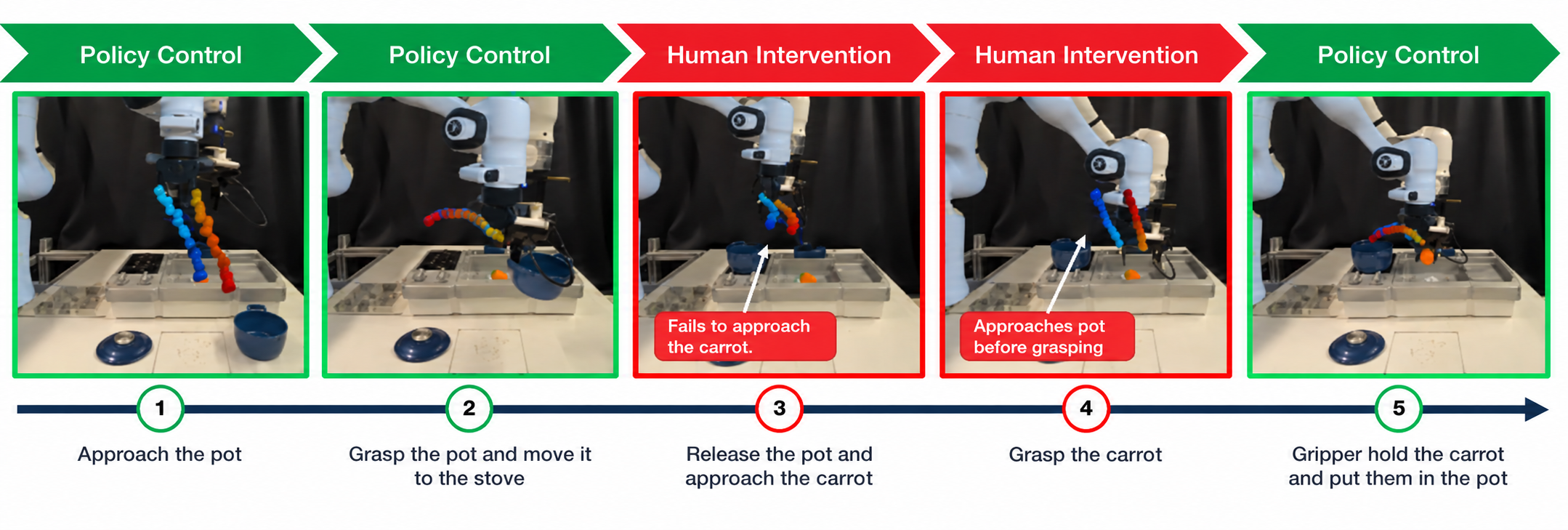}
    \caption{Example of proactive human intervention with PHIL during the Cook the Carrot task.
    Colored spheres visualize the policy-predicted future end-effector trajectories in XR; green and red frames indicate policy control and human intervention, respectively.
    (1--2) The policy approaches the pot and moves it onto the stove.
    (3--4) The operator intervenes when the previewed motion fails to approach the carrot or moves toward the pot before grasping the carrot, providing corrections before these undesirable motions are executed.
    Policy predictions remain visible during human intervention, allowing the operator to assess when to return control.
    (5) Once the predicted motion is acceptable, policy control resumes to place the grasped carrot into the pot.}
    \label{fig:phil_example}
    \PHILfigurebottomspace
\end{figure*}

During policy control, the system receives policy actions of the form $(\mathbf{p}_t,\mathbf{r}_t,g_t)$.
All policies used in our experiments output an absolute end-effector pose in Cartesian space, specified by position $\mathbf{p}_t$ and orientation $\mathbf{r}_t$, together with a binary gripper command $g_t$.
Other action representations can also be used after conversion to absolute end-effector positions and orientations for execution and visualization.
In our experiments, actions are executed at $20\,\mathrm{Hz}$, with an action-chunk horizon of $H=30$ and a query interval of $B=10$ control steps.
Policy queries follow this fixed schedule during both policy control and human intervention.
Each query refreshes the full $H$-step XR preview.
When a new action chunk is received, policy execution switches to the new chunk, discarding any unexecuted actions from the previous chunk.
Fig.~\ref{fig:phil_example} illustrates how PHIL collects proactive correction data in the physical workspace.

\subsection{3D Trajectory Preview via Extended Reality}

The PHIL XR interface runs on the Meta Quest~3 headset using IRIS~\cite{jiang2025iris}.
It displays a virtual Franka Panda whose joint configuration is synchronized with the real robot, helping the operator perceive the robot's current configuration and motion.
QR-code registration aligns the virtual and physical coordinate frames, placing the virtual robot and the policy's predicted trajectories in the same spatial reference as the physical workspace.
The policy's predicted end-effector poses are visualized as a sequence of paired gripper tip points.
Absolute Cartesian poses are rendered directly.
The interface also supports relative Cartesian and joint-space actions by converting them to absolute end-effector poses.
Each pair represents the two gripper tips: their positions and relative arrangement convey the end-effector position and orientation, while their separation indicates whether the gripper is open or closed.
The two tips trace blue and red trajectories that darken toward later predicted actions, allowing the operator to follow the predicted changes in end-effector pose and gripper state over the action chunk.

\subsection{Proactive Human-in-the-loop Robot Behavior Correction}

Reactive correction relies on observed robot execution; PHIL additionally lets the operator anticipate errors through an XR preview.
Policy control continues while the preview is acceptable.
If an error is anticipated, the operator holds the hand trigger to intervene before the undesirable action.
The operator then teleoperates the end effector from the robot's current physical state.
Releasing the trigger immediately restores policy control without waiting for the next query or altering the query schedule.

During human intervention (Fig.~\ref{fig:phil_example}, steps~3--4), policy predictions from the current context continue to update in XR.
For preference-based post-training (Sec.~\ref{sec:proactive_collection}), each rejected prediction $A_t^-$ is paired with the next $H$ executed actions $A_t^+$ from the same query state.
Once the displayed motion is judged acceptable, the operator returns control to the policy (step~5).
The latest displayed prediction is then treated as accepted and excluded from negative samples.

\section{Experiments}
\label{sec:experiments}

\providecommand{\tbd}{\textcolor{red}{\textbf{TBD}}}

This section evaluates PHIL on real-world manipulation tasks to assess its overall performance, the contributions of correction data and preference post-training, and its applicability to a pretrained vision-language-action (VLA) policy.

\subsection{Experimental Setup}

\subsubsection{Tasks}

The experiments cover four real-world robot manipulation tasks.
\textit{Pick Up Banana} and \textit{Towel Folding} are single-stage tasks.
\textit{Cook the Carrot} requires placing a pot on the stove, transferring a carrot into the pot, and closing the lid.
\textit{Lemon in the Drawer} requires opening a drawer, placing a lemon inside, and closing the drawer.
\textit{Cook the Carrot} and \textit{Lemon in the Drawer} each contain three ordered stages.

\subsubsection{Policies}

The experiments use two policy architectures: a diffusion policy and a pretrained vision-language-action (VLA) policy.
The diffusion policy is based on BESO~\cite{reuss2023goal} and is referred to as Diffusion Policy throughout the following experiments.
The VLA policy uses X-VLA~\cite{zheng2025x}.
Diffusion Policy is evaluated on all four tasks.
X-VLA is evaluated on \textit{Towel Folding}, \textit{Cook the Carrot}, and \textit{Lemon in the Drawer}.
\textit{Pick Up Banana} is omitted for X-VLA because the base policy is already close to the performance ceiling.

\subsubsection{Data Budgets}

Base-demonstration budgets vary with task difficulty.
On the simpler \textit{Pick Up Banana} task, Diffusion Policy performance is already close to saturation with 10, 20, or 30 base demonstrations, making gains from additional demonstrations difficult to distinguish.
This task therefore uses three, six, and nine base demonstrations to examine performance changes in a lower-data regime.
Ten, twenty, and thirty base demonstrations are used for the remaining Diffusion Policy tasks and for all X-VLA tasks.

\subsubsection{Metrics}

Full-task success rate (SR) measures the proportion of evaluation rollouts that complete the entire task and is reported for all tasks.
The normalized task completion score (TCS) is additionally reported for the two long-horizon tasks to capture partial completion.
For a task with $K=3$ ordered stages, stage $m$ is credited only when all preceding stages have been completed.
For each rollout, the TCS is defined as $\mathrm{TCS} = \frac{1}{K}\sum_{m=1}^{K}\mathbb{I}_m$, where $\mathbb{I}_m\in\{0,1\}$ indicates whether stage $m$ is completed under this ordering constraint.
SR and TCS are computed over 25 evaluation rollouts per condition.
SR and the mean TCS across evaluation rollouts are reported as percentages.

\subsubsection{Compared Interactive Learning Methods}

The human-in-the-loop imitation learning experiments compare four approaches: Base Policy, HG-DAgger, Residual Policy, and PHIL.
Base Policy serves as the reference policy and is trained only on the base demonstrations.
\textbf{HG-DAgger}~\cite{hg-dagger} post-trains the base policy by behavior cloning with executed corrective trajectories.
\textbf{Residual Policy}, motivated by residual learning from human corrections~\cite{xu2025crdagger}, predicts the action error between a rejected trajectory and the corresponding executed corrective trajectory.
The predicted action error is used as an additive correction to the base policy action.
\textbf{PHIL} post-trains the policy with a pairwise preference objective on rejected--corrected trajectory pairs.

\subsubsection{Corrective Data Collection}

For each policy model, task, and base-demonstration budget, a base policy is first trained on the corresponding demonstrations.
With this base policy held fixed, five reactive and five proactive correction rollout episodes are then collected, each allowing multiple interventions.
The experiments use two corrective data collection modes: reactive and proactive.
During reactive collection, the operator observes the robot's motion directly without the XR headset and intervenes when an error occurs or appears likely.
During proactive collection, the operator uses the XR headset to preview the policy's predicted future trajectory and intervene before an anticipated error occurs.

To support fair comparisons, all interactive learning methods evaluated under the same condition and collection mode reuse the same recorded correction data for adaptation, avoiding differences due to independent data collection.

\subsection{Main Results: Comparison with Interactive Policy Adaptation Baselines}

\begin{figure*}[!t]
    \centering
    \includegraphics[width=\textwidth]{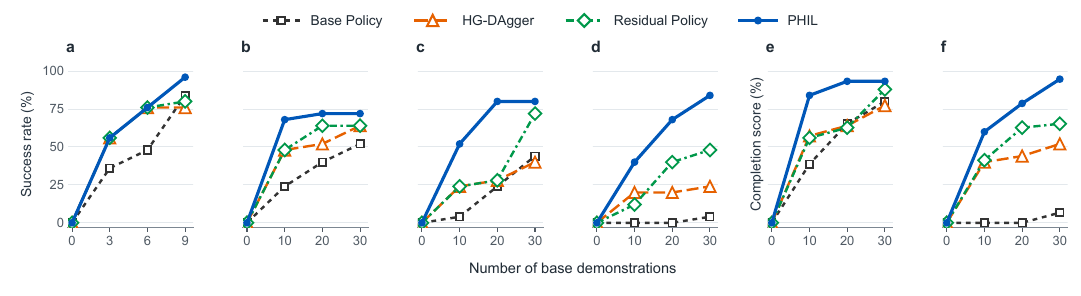}
    \caption{Diffusion Policy performance across base-demonstration budgets.
    Panels (a--d) show full-task success rates for Pick Up Banana, Towel Folding, Cook the Carrot, and Lemon in the Drawer, respectively; panels (e, f) show normalized task completion scores for Cook the Carrot and Lemon in the Drawer.
    Each measured point summarizes 25 evaluation rollouts.
    The points at $(0,0)$ represent an assumed zero-demonstration baseline and were not evaluated experimentally.
    HG-DAgger and Residual Policy use reactive corrections, whereas PHIL uses proactive corrections; all three train with five correction episodes and the original base demonstrations.
    }
    \label{fig:beso_main_results_lineplot}
    \PHILfigurebottomspace
\end{figure*}

Fig.~\ref{fig:beso_main_results_lineplot} compares PHIL with Base Diffusion Policy, HG-DAgger, and Residual Policy across three base-demonstration budgets.
PHIL uses proactive corrections, whereas HG-DAgger and Residual Policy use reactive corrections.
PHIL matches or exceeds both adaptation baselines across all evaluated tasks, budgets, and metrics, with clear gains on the long-horizon tasks.
On \textit{Cook the Carrot} with ten base demonstrations, PHIL achieves 52\% SR and 84.0\% TCS, compared with 4\% SR and 38.7\% TCS for Base Policy, 24\% SR and 57.3\% TCS for HG-DAgger, and 24\% SR and 56.0\% TCS for Residual Policy.

\subsection{Ablations of Correction Collection and Post-Training}

\begin{figure}[!t]
    \centering
    \includegraphics[width=\columnwidth]{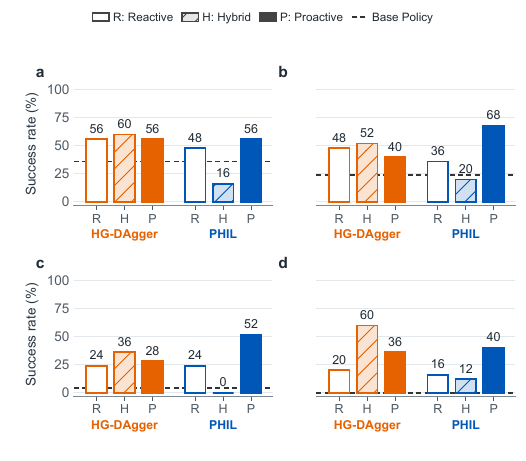}
    \caption{Correction-collection and post-training ablations with Diffusion Policy.
    Panels (a--d) show success rates for Pick Up Banana, Towel Folding, Cook the Carrot, and Lemon in the Drawer, respectively.
    Base policies use three demonstrations for Pick Up Banana and ten for each remaining task.
    Each bar summarizes 25 evaluation rollouts after post-training with five correction episodes and the original base demonstrations.
    R, H, and P denote Reactive, Hybrid, and Proactive collection; Hybrid combines two proactive and three reactive episodes.
    Methods within each collection mode share the same recorded correction data.
    Dashed horizontal lines indicate Base Policy success rates.}
    \label{fig:reactive_proactive_barplot}
    \PHILfigurebottomspace
\end{figure}

Since the main comparison varies both correction collection mode and post-training objective, the following ablations examine their individual contributions.
Fig.~\ref{fig:reactive_proactive_barplot} reports the ablations in the lowest-data regime: three base demonstrations for \textit{Pick Up Banana} and ten for each remaining task.
All adapted conditions start from the corresponding base policy and use five correction episodes, with recorded data shared across methods within each collection mode.

\noindent\textbf{Q1: Does proactive collection improve success when corrections are learned by behavior cloning?}
HG-DAgger is compared across reactive and proactive collection while keeping the behavior-cloning objective fixed.
In both conditions, the correction-data objective learns from positive corrective trajectories without using rejected policy predictions as negative supervision.
Both conditions achieve 56\% SR on \textit{Pick Up Banana}.
Proactive collection yields 40\% SR on \textit{Towel Folding}, compared with 48\% for reactive collection, while increasing SR from 24\% to 28\% on \textit{Cook the Carrot} and from 20\% to 36\% on \textit{Lemon in the Drawer}.
Thus, proactive collection alone does not provide a consistent SR advantage under behavior cloning.

\noindent\textbf{Q2: Does preference post-training improve success on the same proactive correction data?}
HG-DAgger and PHIL use the same proactive positive samples $(c_t,A_t^+)$ extracted from the shared proactive correction data, together with the same original base demonstrations.
HG-DAgger uses behavior cloning to reproduce these positive action chunks, without using the corresponding rejected predictions.
PHIL pairs each positive chunk with its rejected prediction $A_t^-$ and applies a pairwise preference objective that favors the corrective chunk over the rejected prediction.
PHIL matches HG-DAgger at 56\% SR on \textit{Pick Up Banana} and increases SR from 40\% to 68\% on \textit{Towel Folding}, from 28\% to 52\% on \textit{Cook the Carrot}, and from 36\% to 40\% on \textit{Lemon in the Drawer}.
The largest observed gains occur on \textit{Towel Folding} and \textit{Cook the Carrot}, supporting the value of preference post-training with proactive data in these tasks.

\noindent\textbf{Q3: Does proactive collection improve success when the preference objective is fixed?}
In this experimental setting, PHIL learns from positive and negative trajectory pairs, so the way corrections are collected may affect the quality of its training supervision.
To assess whether proactive collection improves task success, PHIL is compared using reactive and proactive correction data under the same pairwise preference objective.
Proactive PHIL achieves higher SR on all four tasks, increasing SR from 48\% to 56\% on \textit{Pick Up Banana}, from 36\% to 68\% on \textit{Towel Folding}, from 24\% to 52\% on \textit{Cook the Carrot}, and from 16\% to 40\% on \textit{Lemon in the Drawer}.
These results support the value of proactively collected data for preference post-training under the evaluated protocol.

The Hybrid conditions combine two proactive and three reactive episodes from the shared correction datasets to examine whether mixing the two collection modes benefits either post-training objective.
HG-DAgger with Hybrid data achieves 60\%, 52\%, 36\%, and 60\% SR on \textit{Pick Up Banana}, \textit{Towel Folding}, \textit{Cook the Carrot}, and \textit{Lemon in the Drawer}, respectively, exceeding its results with either collection mode alone on all four tasks.
In contrast, PHIL with the same Hybrid data achieves 16\%, 20\%, 0\%, and 12\% SR, below its results with either reactive or proactive data alone on every task.
The benefit of mixing the two data sources therefore depends on the post-training objective in these experiments.

\subsection{Recovery Behavior Analysis}

Unnecessary recovery behaviors are defined as recovery-like actions or segments performed when no execution error is present and the task state does not require recovery; necessary recovery after actual errors is excluded.
During the experiments, such behaviors were frequently observed in policies trained with reactive corrections.
This observation motivated a separate experiment to compare their frequency across correction collection modes and post-training methods.

\begin{figure}[!t]
    \centering
    \includegraphics[width=\columnwidth]{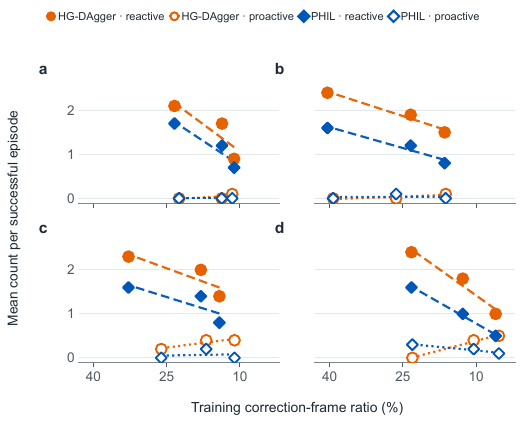}
    \caption{Unnecessary recovery in Diffusion Policies post-trained with HG-DAgger or PHIL.
    Panels (a--d): Pick Up Banana, Towel Folding, Cook the Carrot, and Lemon in the Drawer.
    Each point averages recovery counts over ten successful episodes per method, collection mode, and base-demonstration budget, excluding recovery required after actual errors.
    The x-axis shows correction frames as a percentage of base-demonstration plus correction frames, decreasing rightward.
    Orange circles/blue diamonds denote HG-DAgger/PHIL; filled/open markers denote reactive/proactive collection.
    Dashed/dotted lines show least-squares linear fits across three budgets per method for reactive/proactive collection, respectively.}
    \label{fig:recovery_correction_relationship}
    \PHILfigurebottomspace
\end{figure}

The experiment evaluates Diffusion Policies post-trained with HG-DAgger or PHIL using reactive or proactive data, with ten successful episodes collected per method, collection mode, task, and base-demonstration budget.
Fig.~\ref{fig:recovery_correction_relationship} reports mean counts per episode.
For both methods, proactive-trained policies exhibit fewer such behaviors in every task--demonstration setting, consistent with corrections that more directly describe desired task actions.
This analysis characterizes recovery during successful task completion.

\subsection{PHIL with a Pretrained VLA Policy}

\begin{table}[!t]
    \centering
    \caption{X-VLA results with different numbers of base demonstrations. HG-DAgger and PHIL use corrections collected over five interactive rollout episodes.}
    \label{tab:xvla_comparison}
    \setlength{\tabcolsep}{2.5pt}
    \renewcommand{\arraystretch}{1.16}
    \scriptsize
    \begin{tabular*}{\columnwidth}{@{\extracolsep{\fill}}ll*{6}{r}@{\hspace{3pt}}}
        \toprule
        \multirow{3}{*}{Task} & \multirow{3}{*}{Method}
            & \multicolumn{6}{c}{Base demonstrations} \\
        \cmidrule(l){3-8}
            & & \multicolumn{2}{c}{10} & \multicolumn{2}{c}{20} & \multicolumn{2}{c}{30} \\
        \cmidrule(lr){3-4}\cmidrule(lr){5-6}\cmidrule(l){7-8}
            & & SR & TCS & SR & TCS & SR & TCS \\
        \midrule
        \multirow{3}{*}{\shortstack[l]{Towel\\Folding}}
            & Base Policy & 32 & -- & 36 & -- & 60 & -- \\
            & HG-DAgger & 32 & -- & 40 & -- & 64 & -- \\
            & \textbf{PHIL} & \textbf{64} & -- & \textbf{72} & -- & \textbf{72} & -- \\
        \midrule
        \multirow{3}{*}{\shortstack[l]{Cook the\\Carrot}}
            & Base Policy & 40 & 73.3 & 56 & 77.3 & 56 & 85.3 \\
            & HG-DAgger & \textbf{56} & 78.7 & 48 & 76.0 & \textbf{76} & 90.7 \\
            & \textbf{PHIL} & \textbf{56} & \textbf{85.3} & \textbf{68} & \textbf{89.3} & \textbf{76} & \textbf{92.0} \\
        \midrule
        \multirow{3}{*}{\shortstack[l]{Lemon in\\the Drawer}}
            & Base Policy & 0 & 37.3 & 4 & 46.7 & 0 & 53.3 \\
            & HG-DAgger & 68 & 72.0 & 56 & 78.7 & 80 & 88.0 \\
            & \textbf{PHIL} & \textbf{92} & \textbf{96.0} & \textbf{92} & \textbf{97.3} & \textbf{100} & \textbf{100.0} \\
        \bottomrule
    \end{tabular*}
    \par\vspace{3pt}
    \parbox{\columnwidth}{\scriptsize\raggedright
        All values are percentages over 25 evaluation rollouts.
        SR: full-task success rate; TCS: normalized task completion score.
        --: TCS is not reported for Towel Folding.
        Bold values indicate the best result within each task and demonstration budget, including ties.}
    \PHILfigurebottomspace
\end{table}

Tab.~\ref{tab:xvla_comparison} compares Base Policy, HG-DAgger with reactive corrections, and PHIL with proactive corrections on X-VLA under the shared protocol.
PHIL matches or exceeds HG-DAgger across all evaluated tasks, demonstration budgets, and metrics.
On \textit{Cook the Carrot} with 20 demonstrations, PHIL achieves 68\% SR and 89.3\% TCS, compared with 48\% SR and 76.0\% TCS for HG-DAgger.
With 10 and 30 demonstrations, PHIL matches HG-DAgger's SR while achieving higher TCS.
On \textit{Lemon in the Drawer}, PHIL improves both metrics at every budget, reaching 100\% SR and 100.0\% TCS with 30 demonstrations.
These results support PHIL's applicability to post-training X-VLA on the evaluated tasks.

\section{Conclusion}

We presented PHIL, a method for proactive human-in-the-loop post-training of action-chunking robot policies.
PHIL allows an operator to inspect the policy's upcoming motion during execution and intervene before an anticipated undesirable action occurs.
During each intervention, PHIL pairs retained unexecuted policy predictions with subsequent executed corrective robot action chunks starting from the same query state, allowing the corrective window to extend beyond release.
PHIL trains the policy on these context-aligned pairs using a pairwise preference objective that favors executed corrective trajectories over rejected policy predictions.
Across the evaluated real-world manipulation tasks, policies post-trained with proactive corrections exhibit substantially fewer unnecessary recovery behaviors than those post-trained with reactive corrections.
PHIL matches or exceeds the performance of the compared methods on the evaluated Diffusion Policy tasks and is also applicable to post-training the pretrained X-VLA policy.
These results suggest that pairing unexecuted policy predictions with executed corrective trajectories from the same physical state provides effective supervision for robot policy post-training.

\section*{Acknowledgment}

We used ChatGPT to polish the text of this paper and to generate some concept illustrations in Figs.~\ref{fig:teaser}, \ref{fig:proactive_correction}, and \ref{fig:proactive_preference_pairs}. The authors
reviewed every generated item, and all technical claims, experiments, and
results are our own.

\bibliographystyle{IEEETran}
\bibliography{references}

\clearpage
\appendix


\subsection{Policy Action Representations}
\label{app:policy_action_representations}

In our experiments, BESO and X-VLA use the same action representation: an absolute end-effector position, an absolute Euler-angle orientation, and a binary gripper command.
X-VLA's action representation is adapted to use Euler-angle orientations in this shared action space.

\subsection{Proactive Collection and Preference Post-Training}
\label{app:phil_algorithm}

Algorithm~\ref{alg:hitl_il} summarizes the proactive collection and post-training procedure used in our experiments.
The base policy remains fixed throughout collection, and post-training begins after the rollout data have been collected.
Policy queries follow the same fixed schedule during policy control and human intervention, without interruption or reset at takeover or release.
When the operator releases the trigger after inspecting the most recent displayed prediction, that prediction is labeled accepted and excluded from the negative samples.
Release indicates that the operator judges the displayed trajectory acceptable, including when release occurs between scheduled queries.
The robot then returns to policy control, including the interval before the next scheduled query.

A candidate pair is retained if its shadow prediction was collected during human intervention, is not accepted at release, and all $H$ subsequent executed actions are available in the recorded rollout.
The positive window may extend beyond release and include policy-controlled actions.
Thus, a takeover shorter than $H$ steps can contribute a complete pair.
An intervention contributes no pairs if no scheduled query occurs during human intervention or if none of its predictions satisfies these criteria, including when its only prediction is accepted at release.

\begin{algorithm}[!ht]
\caption{Proactive Correction and Preference Post-Training}
\label{alg:hitl_il}
\small
\begin{algorithmic}[1]
\REQUIRE Base policy $\pi_{\theta_0}$, demonstrations $\mathcal D_{\mathrm{base}}$, horizon $H$, query interval $B\leq H$, rollout budget $E$
\ENSURE Post-trained policy $\pi_\theta$
\STATE Initialize $\mathcal D_{\mathrm{pref}}\leftarrow\emptyset$
\FOR{$e=1$ to $E$}
    \STATE Collect a rollout with $\pi_{\theta_0}$, XR preview, and human takeover; keep $\theta_0$ fixed
    \STATE During the rollout, query every $B$ control steps without resetting the schedule at takeover or release
    \STATE Record each query context $c_t$ and prediction $A_t$, and human intervention intervals
    \STATE At each release, label the most recent displayed prediction as accepted and return to policy control
    \STATE Record the executed actions $(\tilde a_0,\ldots,\tilde a_{T_e-1})$ during both policy control and human intervention
    \FOR{each scheduled query time $t$ during human intervention}
        \IF{$A_t$ is not accepted at release and $t+H\leq T_e$}
            \STATE Set $A_t^-\leftarrow A_t$
            \STATE Set $A_t^+\leftarrow(\tilde a_t,\ldots,\tilde a_{t+H-1})$
            \STATE Append $(c_t,A_t^+,A_t^-)$ to $\mathcal D_{\mathrm{pref}}$
        \ENDIF
    \ENDFOR
\ENDFOR
\STATE Initialize $\theta\leftarrow\theta_0$
\STATE Post-train $\theta$ on $\mathcal D_{\mathrm{base}}$ and $\mathcal D_{\mathrm{pref}}$ using Eq.~\eqref{eq:total_loss}
\RETURN $\pi_\theta$
\end{algorithmic}
\end{algorithm}

\subsection{Supplementary Experimental Details}
\label{app:experimental_details}

This appendix provides the tabulated results and protocol details that complement the figures and analyses in the main paper.
The task-specific demonstration budgets, evaluation metrics, and rollout counts are summarized in Tab.~\ref{tab:experimental_setup}.
SR and TCS are computed over 25 evaluation rollouts per condition, while recovery behaviors are measured in a separate experiment using ten successful episodes per condition.
All adapted methods in the main comparisons and ablations train on data from five correction episodes together with the original base demonstrations to mitigate forgetting.

\begin{table}[!t]
    \centering
    \caption{Task, demonstration-count, and evaluation summary.
    Demo counts refer to base-policy demonstrations.}
    \label{tab:experimental_setup}
    \setlength{\tabcolsep}{2.5pt}
    \renewcommand{\arraystretch}{1.08}
    \scriptsize
    \resizebox{\columnwidth}{!}{%
        \begin{tabular}{lccccc}
            \toprule
            Task & Stages & Diffusion Policy demos & X-VLA demos & Metrics & Rollouts \\
            \midrule
            Pick Up Banana & 1 & 3/6/9 & --- & SR & 25 \\
            Towel Folding & 1 & 10/20/30 & 10/20/30 & SR & 25 \\
            Cook the Carrot & 3 & 10/20/30 & 10/20/30 & SR, TCS & 25 \\
            Lemon in the Drawer & 3 & 10/20/30 & 10/20/30 & SR, TCS & 25 \\
            \bottomrule
        \end{tabular}%
    }
    \vspace{2pt}
    \parbox{\columnwidth}{\scriptsize SR and TCS are computed over 25 evaluation rollouts per condition.
    For post-training, each adaptation condition combines the corresponding base demonstrations with five correction episodes.
    Banana--X-VLA is omitted because its base policy is near the performance ceiling.}
\end{table}

\subsection{Interactive Adaptation Baselines}
\label{app:adaptation_baselines}

This section details how HG-DAgger and Residual Policy use the recorded human corrections in our experiments.
For a given policy model, task, base-demonstration budget, and collection mode, the methods share the same correction rollouts.
Each adaptation condition uses five correction episodes together with the original base demonstrations.
Both baselines use reactive corrections in the main diffusion-policy comparison: the operator observes the robot's executed behavior without an XR preview and takes over when an error becomes visible or appears likely.
The methods differ in the supervision extracted from these rollouts and in how the learned correction affects the policy output.

\subsubsection{HG-DAgger}
\label{app:hg_dagger}

Our HG-DAgger baseline follows human-gated data aggregation~\cite{hg-dagger}, using human takeover to obtain corrective supervision at states visited during policy execution.
The executed corrective trajectories provide target actions for behavior cloning.
Starting from the corresponding base policy, post-training fits these targets while retaining imitation learning on the original demonstrations.
Let $\mathcal D_{\mathrm{corr}}$ denote the resulting context--correction samples $(c,A^+)$.
Using the native imitation loss $\ell_\theta$ defined in Sec.~\ref{sec:preference_posttraining}, the correction-data term is
\begin{equation}
\mathcal L_{\mathrm{corr}}^{\mathrm{HG}}
=
\mathbb E_{(c,A^+)\sim\mathcal D_{\mathrm{corr}}}
\left[\ell_\theta(A^+,c)\right].
\label{eq:hg_correction_loss}
\end{equation}
Training also includes the base-demonstration imitation term in Eq.~\eqref{eq:base_imitation_loss}.
The learned policy then directly generates action chunks for execution.
Rejected base-policy predictions do not serve as negative targets in this behavior-cloning objective.

In the collection-mode ablations, reactive, proactive, and Hybrid HG-DAgger retain this same learning objective while changing the correction data.
Hybrid uses two proactive and three reactive episodes.
For the controlled comparison between proactive HG-DAgger and proactive PHIL, both methods receive exactly the same positive samples $(c_t,A_t^+)$, with the same query start times and complete $H$-step executed windows.
These windows may include policy-controlled actions after the operator releases control, as described in Appendix~\ref{app:phil_algorithm}.
HG-DAgger learns to reproduce these positive chunks; PHIL additionally uses their corresponding rejected predictions in the pairwise preference objective.
HG-DAgger is also used for the X-VLA comparison, where it post-trains the corresponding VLA base policy on reactive corrections and the original demonstrations.

\subsubsection{Residual Policy}
\label{app:residual_policy}

The Residual Policy baseline learns the action error between a rejected base-policy trajectory and its corresponding executed corrective trajectory.
Learning additive corrections from human feedback is also central to CR-DAgger~\cite{xu2025crdagger}.
Here, Residual Policy is evaluated under the shared collection and training protocol described above.
The rejected prediction supplies the reference action, and the executed correction specifies how that action should change.
This provides supervised targets for predicting an additive correction to the base-policy output.
At execution time, the combination can be written at the action level as
\begin{equation}
\hat a_t = a_t^{\mathrm{base}} + \widehat{\Delta a}_t,
\label{eq:residual_action_composition}
\end{equation}
where $a_t^{\mathrm{base}}$ is the base-policy action and $\widehat{\Delta a}_t$ is the predicted action error.
The resulting $\hat a_t$ is the corrected policy output.
The residual therefore represents a change to the base action, while HG-DAgger learns to generate the corrective action directly through the post-trained policy.

The residual baseline uses the same reactive correction rollouts as HG-DAgger in the main diffusion-policy comparison, and its training also includes the original base demonstrations under the shared protocol.
Although Residual Policy and PHIL both use rejected and executed trajectories, they use this information differently: Residual Policy predicts an action-space correction, whereas PHIL compares the imitation losses of the two trajectories to favor the executed continuation.
The main comparison also changes the collection mode, because PHIL uses proactive corrections.
Consequently, the reported performance differences reflect both the correction data and the adaptation objective; the controlled ablations examine these factors separately.

\subsection{Complete Diffusion-Policy Results}
\label{app:diffusion_results}

Tab.~\ref{tab:beso_main_results} reports the complete numerical results for the diffusion-policy comparison visualized in the main paper.
It includes all four tasks and all three base-demonstration budgets, and additionally reports normalized task completion score for the two long-horizon tasks.

\begin{table*}[!t]
    \centering
    \caption{Main diffusion-policy comparison with different numbers of base demonstrations. Pick Up Banana uses 3/6/9 base demonstrations; all other tasks use 10/20/30. Long-horizon tasks report both full-task success rate (SR) and normalized task completion score (TCS).}
    \label{tab:beso_main_results}
    \setlength{\tabcolsep}{2.2pt}
    \renewcommand{\arraystretch}{1.08}
    \scriptsize
    \resizebox{\textwidth}{!}{%
        \begin{tabular}{l*{18}{c}}
            \toprule
            \multirow{3}{*}{Method}
                & \multicolumn{12}{c}{Success Rate (SR)}
                & \multicolumn{6}{c}{Task Completion Score (TCS)} \\
            \cmidrule(lr){2-13}\cmidrule(lr){14-19}
                & \multicolumn{3}{c}{Banana}
                & \multicolumn{3}{c}{Towel}
                & \multicolumn{3}{c}{Carrot}
                & \multicolumn{3}{c}{Lemon}
                & \multicolumn{3}{c}{Carrot}
                & \multicolumn{3}{c}{Lemon} \\
            \cmidrule(lr){2-4}\cmidrule(lr){5-7}\cmidrule(lr){8-10}
            \cmidrule(lr){11-13}\cmidrule(lr){14-16}\cmidrule(lr){17-19}
                & 3 & 6 & 9
                & 10 & 20 & 30
                & 10 & 20 & 30
                & 10 & 20 & 30
                & 10 & 20 & 30
                & 10 & 20 & 30 \\
            \midrule
            Base Policy
                & 36\% & 48\% & 84\%
                & 24\% & 40\% & 52\%
                & 4\% & 24\% & 44\%
                & 0\% & 0\% & 4\%
                & 38.7\% & 65.3\% & 80.0\%
                & 0.0\% & 0.0\% & 6.7\% \\
            HG-DAgger
                & \textbf{56\%} & \textbf{76\%} & 76\%
                & 48\% & 52\% & 64\%
                & 24\% & 28\% & 40\%
                & 20\% & 20\% & 24\%
                & 57.3\% & 64.0\% & 77.3\%
                & 40.0\% & 44.0\% & 52.0\% \\
            Residual Policy
                & \textbf{56\%} & \textbf{76\%} & 80\%
                & 48\% & 64\% & 64\%
                & 24\% & 28\% & 72\%
                & 12\% & 40\% & 48\%
                & 56.0\% & 62.7\% & 88.0\%
                & 41.3\% & 62.7\% & 65.3\% \\
            PHIL
                & \textbf{56\%} & \textbf{76\%} & \textbf{96\%}
                & \textbf{68\%} & \textbf{72\%} & \textbf{72\%}
                & \textbf{52\%} & \textbf{80\%} & \textbf{80\%}
                & \textbf{40\%} & \textbf{68\%} & \textbf{84\%}
                & \textbf{84.0\%} & \textbf{93.3\%} & \textbf{93.3\%}
                & \textbf{60.0\%} & \textbf{78.7\%} & \textbf{94.7\%} \\
            \bottomrule
        \end{tabular}%
    }
    \vspace{2pt}
    \parbox{\textwidth}{\scriptsize The three subcolumns under each task give the number of base-policy demonstrations. All values are percentages.}
\end{table*}

\subsection{Controlled Collection and Objective Ablations}
\label{app:controlled_ablations}

Tab.~\ref{tab:controlled_ablation} gives the complete success-rate results for the controlled ablations.
These conditions isolate the correction collection mode and post-training objective while using the same lowest-data base-policy settings described in the main paper.

\begin{table}[!t]
    \centering
    \caption{Controlled diffusion-policy ablation in the lowest-data regime. Each comparison changes the correction-collection mode or the post-training objective while holding the remaining protocol fixed.}
    \label{tab:controlled_ablation}
    \setlength{\tabcolsep}{2.5pt}
    \renewcommand{\arraystretch}{1.1}
    \scriptsize
    \resizebox{\columnwidth}{!}{%
        \begin{tabular}{llcccc}
            \toprule
            Objective & Collection
                & Banana
                & Towel
                & Carrot
                & Lemon \\
            \midrule
            Base Policy & --- & 36\% & 24\% & 4\% & 0\% \\
            HG-DAgger & Reactive & 56\% & 48\% & 24\% & 20\% \\
            HG-DAgger & Proactive & 56\% & 40\% & 28\% & 36\% \\
            PHIL & Reactive & 48\% & 36\% & 24\% & 16\% \\
            PHIL & Proactive & 56\% & 68\% & 52\% & 40\% \\
            HG-DAgger & Hybrid & 60\% & 52\% & 36\% & 60\% \\
            PHIL & Hybrid & 16\% & 20\% & 0\% & 12\% \\
            \bottomrule
        \end{tabular}%
    }
    \vspace{2pt}
    \parbox{\columnwidth}{\scriptsize All tasks report success rate (SR). All reported values are percentages.}
\end{table}

\subsection{Recovery-Behavior Statistics}
\label{app:recovery_statistics}

Tab.~\ref{tab:recovery_behavior} provides the per-task and per-demonstration-budget statistics underlying the recovery-behavior analysis.
It reports unnecessary recovery counts over ten successful episodes per condition and the proportion of correction frames in the corresponding post-training data for reactive and proactive HG-DAgger collection.
Across the 12 HG-DAgger task--demonstration settings, reactive-trained policies exhibit 214 unnecessary recovery behaviors in total, compared with 21 for proactive-trained policies.

\begin{table}[!t]
    \centering
    \caption{Unnecessary recovery behaviors in a separate evaluation of diffusion policies post-trained with HG-DAgger, and the corresponding training-data composition.
    The comparison covers four tasks and three base-demonstration budgets, with ten successful episodes per condition.}
    \label{tab:recovery_behavior}
    \setlength{\tabcolsep}{1.2pt}
    \renewcommand{\arraystretch}{1.12}
    \scriptsize
    \begin{tabular}{
        >{\raggedright\arraybackslash}p{0.15\columnwidth}
        *{3}{>{\centering\arraybackslash}p{0.058\columnwidth}} |
        *{3}{>{\centering\arraybackslash}p{0.058\columnwidth}} |
        *{3}{>{\centering\arraybackslash}p{0.058\columnwidth}} |
        *{3}{>{\centering\arraybackslash}p{0.058\columnwidth}}
    }
            \toprule
            \multirow{2}{*}{\shortstack[l]{HG-DAgger /\\collection}}
                & \multicolumn{3}{c}{\shortstack{Pick Up\\Banana}}
                & \multicolumn{3}{c}{\shortstack{Towel\\Folding}}
                & \multicolumn{3}{c}{\shortstack{Cook the\\Carrot}}
                & \multicolumn{3}{c}{\shortstack{Lemon in\\the Drawer}} \\
            \cmidrule(lr){2-4}\cmidrule(lr){5-7}\cmidrule(lr){8-10}\cmidrule(lr){11-13}
                & 3 & 6 & 9
                & 10 & 20 & 30
                & 10 & 20 & 30
                & 10 & 20 & 30 \\
            \midrule
            \multicolumn{13}{l}{\textit{Evaluation recovery count}} \\
            Reactive
                & 21 & 17 & 9
                & 24 & 19 & 15
                & 23 & 20 & 14
                & 24 & 18 & 10 \\
            Proactive
                & 0 & 0 & 1
                & 0 & 0 & 1
                & 2 & 4 & 4
                & 0 & 4 & 5 \\
            \midrule
            \multicolumn{13}{l}{\textit{Training correction-frame ratio (\%)}} \\
            Reactive
                & 23.40 & 13.66 & 11.19
                & 40.44 & 23.39 & 16.43
                & 32.82 & 18.00 & 14.20
                & 23.23 & 12.72 & 5.98 \\
            Proactive
                & 22.46 & 13.73 & 11.57
                & 39.27 & 26.38 & 16.22
                & 26.11 & 16.90 & 11.11
                & 23.04 & 10.48 & 5.32 \\
            \bottomrule
    \end{tabular}
    \vspace{2pt}
    \parbox{\columnwidth}{\scriptsize An unnecessary recovery behavior is a recovery-like action or action segment performed when no execution error is present and the current task state does not require recovery.
    Recovery actions needed after an actual error are excluded.
    Count rows report totals over ten successful episodes collected in a separate recovery-behavior experiment.
    For each collection mode, the training ratio is the percentage of correction frames among all base-demonstration and correction frames used for post-training.}
\end{table}

\end{document}